\documentclass[letterpaper]{article} 
\usepackage{aaai2026}  
\usepackage{times}  
\usepackage{helvet}  
\usepackage{courier}  
\usepackage[hyphens]{url}  
\usepackage{graphicx} 
\usepackage{natbib}  
\usepackage{caption} 
\usepackage{algorithm}
\usepackage{algorithmic}

\usepackage{amsmath, bm}
\usepackage{amssymb}
\usepackage{amsthm}
\usepackage{subcaption}
\usepackage[table]{xcolor}
\usepackage{array}
\usepackage{threeparttable}
\newtheorem{theorem}{Theorem}
\newtheorem{definition}{Definition}
\usepackage{mathtools}
\usepackage{multirow}
\usepackage{booktabs}
\usepackage{tabularx}

\usepackage{newfloat}
\usepackage{listings}
\DeclareCaptionStyle{ruled}{labelfont=normalfont,labelsep=colon,strut=off} 
\floatstyle{ruled}
\newfloat{listing}{tb}{lst}{}
\floatname{listing}{Listing}
\title{CTPD: Cross Tokenizer Preference Distillation}
\author{
    Truong Nguyen\textsuperscript{\rm 1}\equalcontrib, 
    Phi Van Dat\textsuperscript{\rm 1}\equalcontrib, 
    Ngan Nguyen\textsuperscript{\rm 2}\equalcontrib, 
    Linh Ngo Van\textsuperscript{\rm 1}\thanks{Corresponding Author.}, 
    Trung Le\textsuperscript{\rm 3}, 
    Thanh Hong Nguyen\textsuperscript{\rm 4}
}
\affiliations{
    \textsuperscript{\rm 1}Hanoi University of Science and Technology, Hanoi, Vietnam\\
    \textsuperscript{\rm 2}FPT Smart Cloud, Hanoi, Vietnam\\
    \textsuperscript{\rm 3} Monash University, Clayton, VIC 3800, Australia\\
    \textsuperscript{\rm 4}University of Oregon Eugene, Oregon, United States\\
    truong.nd235566@sis.hust.edu.vn,
    dat.pv235034@sis.hust.edu.vn,
    ngannt61@fpt.com,
    linhnv@soict.hust.edu.vn, trunglm@monash.edu,
    thanhhng@cs.uoregon.edu
}

\begin{document}

\maketitle

\begin{abstract}
While knowledge distillation has seen widespread use in pre-training and instruction tuning, its application to aligning language models with human preferences remains underexplored, particularly in the more realistic cross-tokenizer setting. The incompatibility of tokenization schemes between teacher and student models has largely prevented fine-grained, white-box distillation of preference information. To address this gap, we propose Cross-Tokenizer Preference Distillation (CTPD), the first unified framework for transferring human-aligned behavior between models with heterogeneous tokenizers. CTPD introduces three key innovations: (1) Aligned Span Projection, which maps teacher and student tokens to shared character-level spans for precise supervision transfer; (2) a cross-tokenizer adaptation of Token-level Importance Sampling (TIS-DPO) for improved credit assignment; and (3) a Teacher-Anchored Reference, allowing the student to directly leverage the teacher’s preferences in a DPO-style objective. Our theoretical analysis grounds CTPD in importance sampling, and experiments across multiple benchmarks confirm its effectiveness, with significant performance gains over existing methods. These results establish CTPD as a practical and general solution for preference distillation across diverse tokenization schemes, opening the door to more accessible and efficient alignment of language models.
\end{abstract}

\begin{links}
    \link{Code}{https://github.com/dinhtruongng/CTPD}
    
\end{links}

\section{Introduction}
Aligning Large Language Models (LLMs) with human values and preferences has become a cornerstone of modern AI research. This alignment aims to guide LLMs to generated outputs that are not only fluent but also beneficial, non-harmful, and consistent with intricate human norms. While early efforts relied on Reinforcement Learning from Human Feedback (RLHF)~\cite{christiano2017deep}, recent methods like Direct Preference Optimization (DPO)~\cite{rafailov2023direct} and its variants offer more stable and computationally efficient alternatives, proving highly effective in creating state-of-the-art, user-aligned models. The effectiveness of preference alignment has been primarily demonstrated on large-scale, proprietary language models. However, the substantial computational requirements and the closed-source nature of these models pose significant barriers to accessibility and broad adoption, particularly in resource-constrained settings. In contrast, small language models (SLMs) offer a more practical alternative in such contexts but face notable challenges in achieving alignment comparable to that of larger models, largely due to their limited representational capacity. This often leads to an \textbf{alignment tax} after RLHF training, where their broad task performance is negatively impacted~\cite{bai2022training}.

Knowledge distillation (KD)~\cite{hinton2015distilling} offers a promising solution, where a smaller student model learns from a larger, pre-aligned teacher. This approach is efficient, as the costly alignment process is performed only once by the teacher. While black-box KD methods use only teacher output text, white-box methods leverage richer internal signals like logits for more fine-grained supervision. However, white-box distillation faces a critical obstacle: the cross-tokenizer problem. Teacher and student models often use different tokenizers, leading to incompatible logit distributions and preventing direct token-level knowledge transfer.

Although knowledge distillation has been extensively studied in the contexts of pre-training and instruction tuning~\cite{zhang2024dual, boizard2024towards, cui2025multi}, its application to the critical task of aligning language models with human preferences remains relatively underexplored. To date, only a single work~\cite{gao2025adpa} has investigated white-box distillation in this setting, and it was restricted to a simplified scenario where the teacher and student share an identical tokenizer. Importantly, the more realistic and challenging case of cross-tokenizer distillation for preference alignment has received little to no attention in the literature. Given the abundance of high-performing large language models (LLMs) with varying architectures and tokenization schemes that could serve as teacher models, advancing cross-tokenizer distillation techniques for human preference alignment is crucial to fully exploit their capabilities. However, existing approaches designed for cross-tokenizer distillation in pretraining or finetuning \cite{zhang2024dual,boizard2024towards,cui2025multi} are not directly applicable to this setting. These methods are primarily tailored to align the final-layer logits of teacher and student models for general-purpose learning tasks and do not address the specific challenges posed by preference-based supervision.

To bridge this gap, we propose Cross-Tokenizer Preference Distillation (CTPD) which is the first unified framework that enables the transfer of human-aligned behavior from a high-capacity teacher model to a smaller student model. CTPD is motivated by the observation that, while the tokenizations used by teacher and student models may differ syntactically, both ultimately encode the same underlying natural language substrings. By projecting the teacher’s supervision signals onto the student’s tokens through precisely aligned character-level spans and redefining the DPO objective accordingly, CTPD enables fine-grained white-box supervision even in the presence of heterogeneous tokenizers. Concretely, CTPD comprises three key components:
\begin{enumerate}
  \item Aligned Span Projection: CTPD constructs a dynamic lattice to partition input sequences into aligned spans—pairs of teacher and student token subsequences that correspond to identical character-level intervals. This alignment allows us to compute projected log-probabilities over the student vocabulary without introducing any additional learnable parameters.
  \item Cross-tokenizer Importance Weighting: Building on this alignment, we extend the TIS-DPO framework \cite{liu2024tisdpo} to the cross-tokenizer setting. Token-level importance weights from the teacher are aggregated within each aligned span and transferred to the corresponding student spans, resulting in span-specific weights that enhance credit assignment across mismatched token spaces.
  \item Teacher-Anchored Reference: CTPD adopts the teacher model itself as the reference distribution $\pi_{\text{ref}}$ in the DPO-style objective. Through the span projection mechanism, the student can approximate the teacher’s log-probabilities over its own tokens, enabling the definition of a teacher-anchored DPO-style objective. This loss function retains the structure of standard DPO but naturally accommodates heterogeneous tokenizers, allowing the student to benefit directly from the teacher’s preferences.
\end{enumerate}

CTPD addresses the core problem cross-tokenizer in preference distillation, providing the first practical solution for full-resolution white-box preference transfer. By decoupling alignment from tokenizer compatibility, CTPD makes it feasible to distill sophisticated alignment behaviors from any powerful teacher into any smaller student, thereby facilitating the development of efficient and robustly aligned language models. Especially, we provide a theoretical foundation for the CTPD framework based on importance sampling, which enhances its reliability and provides deeper insights into the dynamics of cross-tokenizer preference distillation.

We conduct extensive experiments to demonstrate significant improvements of CTPD across multiple benchmarks. Furthermore, comprehensive ablation study and analysis confirm the effectiveness of our weighting strategy, the aligned span and teacher-anchored approaches, providing valuable insights into the underexplored space of preference distillation.

\section{Related work}
\subsection{Preference Alignment}

The prevailing approach for human alignment is Reinforcement Learning from Human Feedback (RLHF)~\cite{christiano2017deep, stiennon2020learning, ouyang2022training}. This multi-stage process, which involves training a reward model and then optimizing a policy with reinforcement learning (e.g., PPO~\cite{schulman2017proximal}), has shown empirical success but is often criticized for its training complexity and instability~\cite{rafailov2023direct, bai2022training, gpt4}. To mitigate these issues, Direct Preference Optimization (DPO)~\cite{rafailov2023direct} was introduced as a more direct method that bypasses the explicit reward modeling and RL loop. DPO reframes the problem as a simple binary classification task on preference pairs, enabling stable training via a simple objective and demonstrating performance competitive with PPO-based RLHF~\cite{rafailov2023direct}.

Building on DPO's success, several extensions have emerged. Of particular relevance to our work is Token-level Importance-Sampling DPO (TIS-DPO)~\cite{liu2024tisdpo}, which addresses DPO's uniform treatment of all tokens in a sequence. By introducing token-level importance weights, TIS-DPO concentrates the learning signal on the most salient tokens, improving credit assignment and alignment efficiency. We build upon this insight to extend the TIS-DPO framework to the cross-tokenizer distillation setting.

\subsection{Knowledge Distillation}
Knowledge Distillation (KD) is a model compression technique where a compact student model is trained to emulate a larger teacher model, aiming to transfer its knowledge and achieve comparable performance with significantly reduced computational cost~\cite{hinton2015distilling}. KD methodologies are broadly classified into two categories: black-box distillation and white-box distillation.

Black-box distillation uses only the teacher's final text outputs to create synthetic training data, a simple approach used in instruction tuning but which discards the teacher's rich internal knowledge~\cite{hsieh2023distilling}. In contrast, white-box distillation leverages the teacher's internal logits. These \textit{soft targets} provide more fine-grained supervision by capturing the teacher's full probability distribution over its vocabulary, including its confidence and uncertainty.

A key challenge for white-box distillation is the cross-tokenizer problem, which arises when teacher and student models have incompatible vocabularies. While some recent work has addressed this for general-purpose KD~\cite{zhang2024dual, boizard2024towards, cui2025multi, truong-etal-2025-emo}, its application to preference alignment is almost entirely unexplored. To our knowledge, only one study has investigated white-box preference distillation~\cite{gao2025adpa}, and it was limited to a scenario with a shared tokenizer, thereby avoiding the cross-tokenizer challenge that our work directly addresses.

\section{Methodology}
\subsection{Preliminaries}

Reinforcement learning from human feedback (RLHF) \cite{christiano2017deep} typically begins with a preference dataset, denoted as $\mathcal{D}$, which consists of tuples $(x, y_w, y_l)$. In each tuple, $x$ is the input prompt, $y_w$ is the response preferred by humans, and $y_l$ is the dispreferred response. Using this data, a sequence-level reward model (RM) is trained with the following objective:


\begin{align*}
\mathcal{L}_{\text{RM}}
    &= -\mathbb{E}_{(x, y_w, y_l) \sim \mathcal{D}} \\
    &\quad\Big[ \log \sigma\big( \text{RM}_{\phi}(x, y_w) - \text{RM}_{\phi}(x, y_l) \big) \Big]
\end{align*}

where $\sigma$ is the sigmoid function. The policy $\pi_\theta$ is then optimized via techniques like PPO \cite{schulman2017proximal} to maximize the reward from the RM, constrained by a KL-divergence from a reference policy $\pi_{\text{ref}}$:
\begin{equation*}
    \max_{\theta} \mathbb{E}_{y \sim \pi_{\theta}(\cdot|x)} \left[ \text{RM}(x, y) - \beta \log \frac{\pi_{\theta}(y|x)}{\pi_{\text{ref}}(y|x)} \right]
\end{equation*}

where $\pi_{\text{ref}}$ denotes the reference policy.
Direct Preference Optimization (DPO) \cite{rafailov2023direct} bypasses the reward modeling step by directly optimizing the policy on preference pairs using the following loss function:
\begin{equation*} \label{eq:dpo}
\mathcal{L}_{\text{DPO}} = -\mathbb{E}_{(x, y_w, y_l) \sim \mathcal{D}} \left[ \log \sigma \left( \beta (r(x, y_w)-r(x,y_l)) \right) \right]
\end{equation*}

where $r(x, y) = \log \frac{\pi_{\theta}(y|x)}{\pi_{\text{ref}}(y|x)}$. TIS-DPO \cite{liu2024tisdpo} extends DPO by introducing token-level importance weights $w_t$ to re-weight the per-token log-odds, focusing the optimization on the most critical tokens. While the complete objective function includes a sequence KL term, the original paper indicates that this component has a negligible impact on the final outcome. Consequently, the weighted token-level reward can be regarded as the principal element within the objective function:

\begin{multline*}
u(x, y_w, y_l, \pi_\theta, w^w, w^l) = \beta (r(x, y_w) - r(x, y_l))
\end{multline*}

where $r(x, y)=\sum_{i=1}^{T} w_i \log \frac{\pi_\theta(y_i \mid x, y_{<i})}{\pi_{\text{ref}}(y_i \mid x, y_{<i})}$. 

\subsubsection{Discussion: Reference Model as a Reweighting Mechanism.}
\label{teacher_ref}
From the loss function of DPO above, we can derive the gradient with respect to the parameters \( \theta \):

\begin{equation*}
\nabla_\theta \mathcal{L}_{\text{DPO}} 
= -\beta \, \mathbb{E}_{(x, y_w, y_l) \sim \mathcal{D}} \left[ \lambda \cdot \nabla_\theta \log \frac{\pi_\theta(y_w \mid x)}{\pi_\theta(y_l \mid x)} \right]
\end{equation*}

where \( \lambda \) is defined as:

\begin{equation*}
\lambda = \sigma\left( 
\beta \log \frac{\pi_{\text{ref}}(y_w \mid x)}{\pi_{\text{ref}}(y_l \mid x)} 
- \beta \log \frac{\pi_\theta(y_w \mid x)}{\pi_\theta(y_l \mid x)}
\right)
\end{equation*}

From the perspective of example reweighting~\cite{ren2018learning}, DPO learns from preference pairs with weights \( \lambda \), where the reference model \( \pi_{\text{ref}} \) controls the training process by adjusting \( \lambda \).

As training progresses, the reference continuously constrains the policy’s deviation by adjusting the value of \( \lambda \). Specifically, when $\frac{\pi_{\text{ref}}(y_w \mid x)}{\pi_{\text{ref}}(y_l \mid x)}$ is large, it encourages a larger value of \( \lambda \), promoting learning from the corresponding preference pair. A small ratio typically results in a reduced value of \( \lambda \), which can reduce the model’s learning from that sample. Therefore, a suboptimally configured reference model can lead to suboptimal weighting of training samples. This observation suggest that employing a highly capable reference model from the outset would achieve better preference optimization results. Our proposed framework aims to leverage this insight; however, a significant challenge arises from the divergent tokenizers employed by the student and teacher models. This discrepancy makes it impossible to compute the log ratio between the policy and the reference model. To address this limitation, the subsequent section introduces the notion of an \textbf{aligned span}, which serves to connect and align the student and teacher models. This connection thereby enables the distillation of preferences from the teacher reference, overcoming the challenge of incompatible tokenizers.

\subsection{Aligned Span}
The foundation of our cross-tokenizer framework is an alignment mechanism that uses the original, untokenized string as a common ground truth. Instead of relying on heuristics like word boundaries, we align tokens based on the exact character indices they represent in the source string. The objective is to find subsequences of tokens from both the teacher and student models that map to the identical character-level span.

Let $S$ be the original string. Any token $t_i$ from the teacher's tokenizer and $s_j$ from the student's tokenizer corresponds to a specific substring of $S$, which can be identified by its `(start, end)` character indices. Our method partitions the full token sequences into a series of aligned spans.

\begin{definition} \label{definition1}
A teacher token subsequence $\{t_i, \dots, t_j\}$ and a student token subsequence $\{s_k, \dots, s_l\}$ form an \textbf{aligned span} if the union of their decoded characters covers the exact same start and end index in the original string $S$.
\end{definition}

Our framework below partitions the input text into aligned spans and then processes it at the span level. This mechanism allows us to confidently aggregate any signal (e.g., log-probabilities, importance weights) from the teacher tokens within a span and project it onto the corresponding student tokens in the same span. This method guarantees a sound basis for white-box distillation, eliminating any ambiguity or information loss from tokenizer mismatch. The following section details our proposed framework, which is built upon the concept of aligned spans.
\begin{figure*}[t]
  \centering
  \includegraphics[width=0.8\linewidth]{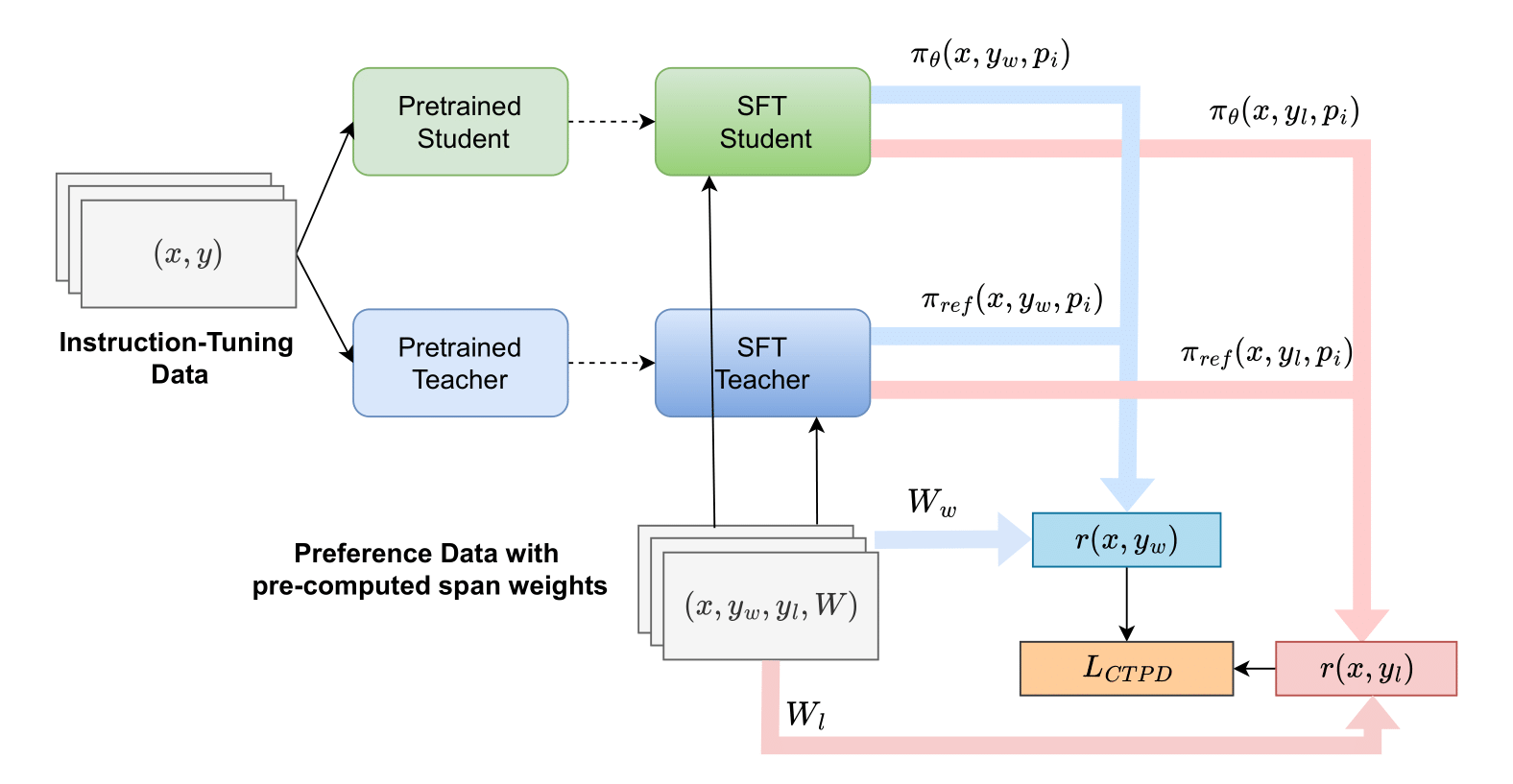}
  \caption{\textbf{An overview of the Cross-Tokenizer Preference Distillation (CTPD) framework.} Initially, both a student and a stronger teacher model are supervised fine-tuned (SFT) using instruction-tuning data. The SFT student model is then further trained using preference data, which consists of winning $y_w$ and losing $y_l$ responses, along with pre-compute aligned span weights. The SFT teacher model serves as a reference to calculate the rewards for aligned spans within these responses. These rewards, along with pre-computed span weights $W$, are ultimately used to compute the objective $L_{CTPD}$, effectively guiding the student model to better align with the preferred outputs.}
  \label{fig:ctpd}
\end{figure*}

\subsection{Cross Tokenizer Preference Distillation Framework}
Recent work on Token-level Direct Preference Optimization (TDPO)~\cite{zeng2024token} establishes that the overall sequence reward can be decomposed into the sum of rewards for individual tokens.
The reward for a given token $y_i$ is defined as below: 
\begin{equation*}
r(y^i | x, y_{<i}) = \beta \log \frac{\pi_\theta(y_{i} \mid x, y_{{<i}})}{\pi_{\text{ref}}(y_{i} \mid x, y_{{<i}})}
\end{equation*}

Assume the probability of an aligned span equals to the product of its corresponding tokens. For an aligned span $p^t$, with corresponding tokens set is $\{y_{t_1}, y_{t_2}, \dots, y_{t_n}\}$, we have:
\begin{equation*}
\pi(p^t \mid x, p^{<t}) = \prod_{i} \pi(y_{t_i} \mid x, y_{t_<i})
\end{equation*}

So the reward of an aligned span would be equal to the sum of its corresponding tokens.
\begin{equation*}
r(p^t | x, p^{<t}) = \sum_{i} r(y_{t_i} | x, y_{t_{<i}})
\end{equation*}

Drawing inspiration from TIS-DPO and applying it to our aligned span structure, we posit that significant fluctuations in these \textbf{span-level} rewards within a response are indicative of label noise in the preference data. The following theorem formalizes this relationship.

\begin{theorem}[Label noise at span level] \label{theorem1}
Let $r_{w,1}, \dots, r_{w,n_w}$ be a set of $n_w$ independent bounded random variables in $[a_w, b_w]$ representing the rewards of the aligned spans in a winning response. Similarly, let $r_{l,1}, \dots, r_{l,n_l}$ be $n_l$ independent bounded random variables in $[a_l, b_l]$ for a losing response. Let their respective average rewards be $S_w = \frac{1}{n_w} \sum_{i=1}^{n_w} r_{w,i}$ and $S_l = \frac{1}{n_l} \sum_{j=1}^{n_l} r_{l,j}$. Then, the probability of the event $S_w \leq S_l$, which signifies data noise, is bounded by:
\begin{equation*} \label{eq:th1_v2}
P(S_w \leq S_l) \leq \exp \left( - \frac{2(\mathbb{E}[S_w] - \mathbb{E}[S_l])^2}{\sum_{i=1}^{n_w} c_{w,i}^2/n_w^2 + \sum_{j=1}^{n_l} c_{l,j}^2/n_l^2} \right)
\end{equation*}
In this expression, $c_{w,i} = b_w - a_w$ and $c_{l,j} = b_l - a_l$ denote the maximum possible change in reward for any single aligned span.
\end{theorem}

To mitigate this noise and promote more stable optimization, we need to ensure consistent rewards for the aligned span $p^t$ across all positions t. Therefore, we define the optimal dataset distribution $D^*$ as follows:

\begin{definition}[Span-level optimal dataset] \label{definition2}
An optimal dataset, denoted by $\mathcal{D}^*$, is characterized by the property that for any given context $(x, p^{<t})$, the subsequent aligned span $p^t$ is drawn from a distribution such that its expected reward is a constant value $R^*$. Formally, for all $(x, p^{<t}) \in \mathcal{D}^*$:
\begin{equation*} \label{eq:7_v1}
\mathbb{E}_{p^t \sim \mathcal{D}^*(\cdot | x, p^{<t})}[r(p^t | x, p^{<t})] = R^*
\end{equation*}
In this expression, $\mathcal{D}^*(\cdot | x, p^{<t})$ represents the conditional probability distribution over the next aligned span $p^t$ given the preceding context, as defined by the optimal dataset.
\end{definition}

Based on Definition \ref{definition2}, we can derive the relationship between the real data $D$ and the optimal data $D^*$ with the following theorem.

\begin{theorem} \label{th2}
Suppose that for an original dataset $\mathcal{D}$, there corresponds an ideal dataset $\mathcal{D}^*$ which satisfies the constant expected reward property outlined in Definition \ref{definition2}. Under this condition, the probability distribution $\mathcal{D}^*(x, p^{<t}, p^t)$ of the ideal dataset is necessarily a re-weighted version of the original distribution $\mathcal{D}$, given by the relation:
\begin{equation*} \label{eq:th2}
\mathcal{D}^*(x, p^{<t}, p^t) = \frac{\mathcal{D}(x, p^{<t}, p^t)}{w(p^t \mid x, p^{<t})}
\end{equation*}
where the weighting function, $w(p^t \mid x, p^{<t})$, is defined as:
\begin{equation*}
w(p^t \mid x, p^{<t}) = k \cdot \exp\left( \mu r(p^t \mid x, p^{<t}) \right)
\end{equation*}
In this formulation, \( p^t \) represents an aligned span, while \(k\) and \(\mu\) are constants that depend on the given context $(x, p^{<t})$.
\end{theorem}

Directly sampling from the ideal distribution \(\mathcal{D}^*\) is intractable in practice. However, the relationship in Theorem 2 frames the problem perfectly for importance sampling~\cite{kloek1978bayesian}. We can sample from our real dataset \(\mathcal{D}\) and use the weights \(w(p^t \mid x, p^{<t})\) to correct for the difference, effectively optimizing on the ideal distribution \(\mathcal{D}^*\).

Inspired by TIS-DPO~\cite{liu2024tisdpo}, we define our primary objective in an idealized setting. Assuming access to the optimal, noise-free dataset $\mathcal{D}^*$ from Definition \ref{definition2}, the Cross-Tokenizer Preference Distillation (CTPD) loss is:
\begin{equation*}
    \mathcal{L}_{\textit{CTPD}} = -\mathbb{E}_{(x, y_w, y_l) \sim \mathcal{D^*}} \left[ \log \sigma \left( \beta(r(x, y_w) 
 - r(x,y_l))\right) \right]
    \label{eq:tis_dpo}
\end{equation*}

where $r(x, y)=\sum_{i=1}^{T} \log \frac{\pi_\theta(p_i \mid x, p_{<i})}{\pi_{\text{ref}}(p_i \mid x, p_{<i})}$ and $p_i$ is the $i_{th}$ aligned span of the sequence $y$. The objective is defined over the ideal dataset $\mathcal{D}^*$, which is not accessible in practice. To formulate a trainable objective using our real dataset $\mathcal{D}$, we employ importance sampling and leverage the relationship in Theorem \ref{th2}. The expected value of reward for an aligned span $p_t$ under $\mathcal{D}^*$, with the form of $r(p_t) = \log \frac{\pi_{\theta}(p_t \mid x,p_{<t})}{\pi_{\text{ref}}(p_t \mid x, p_{<t})}$, can be re-expressed as an unbiased expectation over $\mathcal{D}$:

\begin{equation*} \label{eq:est}
    \mathbb{E}_{x, p_{<t}, p_t \sim D^*} \left(r(p_t)\right) = \mathbb{E}_{x, p_{<t}, p_t \sim D} \left( r(p_t) \cdot w^t \right)
\end{equation*}

with $w^t = \frac{1}{w(p_t \mid x, p_{<t})}$. Using this unbiased estimator as a heuristic and plug it into our main loss yields the final, practical CTPD objective, which is optimized over the real dataset $\mathcal{D}$:

\begin{equation*}
\mathcal{L}_{\text{CTPD}} = - \mathbb{E}_{(x, y_w, y_l) \sim D} \left[ \log \sigma\left( \beta(r(x, y_w) 
 - r(x,y_l)) \right) \right]
\end{equation*}

with $r(x, y)=\sum_{i=1}^{T} w_i \log \frac{\pi_\theta(p_i \mid x, p_{<i})}{\pi_{\text{ref}}(p_i \mid x, p_{<i})}$. Based on the analysis at Section~\ref{teacher_ref}, we will employ a stronger teacher model as a reference model to provides foresight into promising directions for policy improvement base on preference data $D$, allowing for more effective data reweighting and guidance during training. 

All the proofs and derivations could be found in the Extended version of this paper.

\begin{table*}[t]
\centering
\begin{threeparttable}
\centering

\definecolor{lightgray}{gray}{0.9}

\begin{tabular}{lccccccc}
\toprule
\textbf{Method} & \textbf{HellaSwag} & \textbf{Arc} & \textbf{MMLU} & \textbf{TruthfulQA} & \textbf{Winogrande} & \textbf{GSM8k} & \textbf{Average} \\
\midrule
\multicolumn{8}{c}{\cellcolor{lightgray} Qwen-2.5-14B $\rightarrow$ Llama-3.1-8B} \\
\midrule
Teacher & $84.34_{\pm0.36}$ & $67.06_{\pm0.14}$ & $79.74_{\pm0.32}$ & $58.51_{\pm0.15}$ & $80.58_{\pm0.11}$ & $84.23_{\pm0.10}$ & $75.74$ \\
Student & $81.99_{\pm0.38}$ & $57.59_{\pm0.14}$ & $65.48_{\pm0.37}$ & $45.19_{\pm0.14}$ & $77.43_{\pm0.10}$ & $50.27_{\pm0.13}$ & $62.99$ \\
\midrule
SFT & $80.94_{\pm0.39}$ & $60.92_{\pm0.13}$ & $65.58_{\pm0.38}$ & $51.72_{\pm0.14}$ & $77.42_{\pm0.17}$ & $50.64_{\pm0.13}$ & $64.54$ \\
DPO & $\textbf{82.42}_{\pm0.37}$ & $60.84_{\pm0.14}$ & $65.26_{\pm0.38}$ & $52.16_{\pm0.15}$ & $78.31_{\pm0.21}$ & $\underline{54.87}_{\pm0.23}$ & $65.64$ \\
TIS-DPO & $81.08_{\pm0.37}$ & $\underline{61.92}_{\pm0.13}$ & $\textbf{66.73}_{\pm0.31}$ & $\underline{53.86}_{\pm0.10}$ & $\underline{79.05}_{\pm0.12}$ & $54.31_{\pm0.10}$ & $\underline{66.16}$ \\
\midrule
DSKD & $79.24_{\pm0.40}$ & $58.19_{\pm0.12}$ & $64.82_{\pm0.38}$ & $51.77_{\pm0.15}$ & $74.82_{\pm0.21}$ & $50.11_{\pm0.14}$ & $63.16$ \\
ULD & $79.36_{\pm0.39}$ & $57.69_{\pm0.14}$ & $64.96_{\pm0.38}$ & $50.31_{\pm0.18}$ & $77.66_{\pm0.11}$ & $50.16_{\pm0.42}$ & $63.35$ \\
Multi-Level OT & $80.87_{\pm0.39}$ & $60.93_{\pm0.23}$ & $65.39_{\pm0.38}$ & $51.99_{\pm0.18}$ & $77.35_{\pm0.11}$ & $50.95_{\pm0.18}$ & $64.58$ \\
\midrule
\textbf{CTPD (ours)} & $\underline{82.25}_{\pm0.34}$ & $\textbf{63.92}_{\pm0.14}$ & $\underline{66.65}_{\pm0.38}$ & $\textbf{55.22}_{\pm0.15}$ & $\textbf{79.29}_{\pm0.11}$ & $\textbf{57.47}_{\pm0.13}$ & $\textbf{67.42}$ \\
\midrule
\multicolumn{8}{c}{\cellcolor{lightgray} Qwen-2.5-7B $\rightarrow$ Llama-3.2-1B} \\
\midrule
Teacher & $80.34_{\pm0.37}$ & $63.57_{\pm0.15}$ & $74.28_{\pm0.39}$ & $56.37_{\pm0.14}$ & $75.77_{\pm0.11}$ & $81.34_{\pm0.54}$ & $71.95$ \\
Student & $65.59_{\pm0.38}$ & $39.33_{\pm0.16}$ & $31.86_{\pm0.34}$ & $37.66_{\pm0.12}$ & $62.75_{\pm0.12}$ & $6.82_{\pm0.13}$ & $40.67$ \\
\midrule
SFT & $65.95_{\pm0.41}$ & $39.59_{\pm0.14}$ & $\textbf{31.73}_{\pm0.35}$ & $41.17_{\pm0.16}$ & $62.87_{\pm0.15}$ & $6.78_{\pm0.69}$ & 41.35 \\
DPO & $\underline{66.35}_{\pm0.47}$ & $40.10_{\pm0.21}$ & $31.13_{\pm0.38}$ & $41.79_{\pm0.38}$ & $63.30_{\pm0.29}$ & $7.43_{\pm0.72}$ & 41.68 \\
TIS-DPO & $66.23_{\pm0.43}$ & $\textbf{40.92}_{\pm0.15}$ & $\underline{31.43}_{\pm0.37}$ & $\underline{43.49}_{\pm0.14}$ & $\underline{64.34}_{\pm0.13}$ & $\underline{9.13}_{\pm0.71}$ & $\underline{42.60}$ \\
\midrule
DSKD & $65.05_{\pm0.48}$ & $40.16_{\pm0.14}$ & $31.11_{\pm0.38}$ & $40.72_{\pm0.17}$ & $62.89_{\pm0.13}$ & $6.77_{\pm0.56}$ & 41.12 \\
ULD & $65.09_{\pm0.49}$ & $40.02_{\pm0.13}$ & $31.15_{\pm0.38}$ & $41.20_{\pm0.12}$ & $62.77_{\pm0.31}$ & $5.77_{\pm0.63}$ & 41.00 \\
Multi-Level OT & $65.46_{\pm0.42}$ & $39.76_{\pm0.13}$ & $31.19_{\pm0.39}$ & $41.73_{\pm0.23}$ & $63.14_{\pm0.16}$ & $7.12_{\pm0.61}$ & 41.40 \\
\midrule
\textbf{CTPD (ours)} & $\textbf{67.30}_{\pm0.46}$ & $\underline{40.61}_{\pm0.15}$ & $31.08_{\pm0.23}$ & $\textbf{46.34}_{\pm0.14}$ & $\textbf{64.50}_{\pm0.14}$ & $\textbf{9.72}_{\pm0.77}$ & $\textbf{43.26}$ \\
\bottomrule
\end{tabular}

\caption{Benchmark results comparing CTPD and various baselines methods of preference alignment and knowledge distillation. All scores are reported with $\pm$ standard error, computed using the default settings of
\texttt{lm-eval-harness}.}
\label{tab:benchmark_results_booktabs}

\end{threeparttable}
\end{table*}

\begin{table*}[t] 
\centering
\begin{threeparttable}
\centering

\definecolor{lightgray}{gray}{0.9}

\begin{tabular}{lccccccc}
\toprule
\textbf{Method} & \textbf{HellaSwag} & \textbf{Arc} & \textbf{MMLU} & \textbf{TruthfulQA} & \textbf{Winogrande} & \textbf{GSM8k} & \textbf{Average} \\
\midrule
\textbf{Origin} & 82.25 & 63.92 & 66.65 & 55.22 & 79.29 & 57.47 & 67.42 \\
\midrule
Random & 72.13 & 50.45 & 52.11 & 39.26 & 69.58 & 45.28 & 54.80 \\
Average & 81.34 & 59.78 & 65.35 & 53.49 & 77.29 & 55.58 & 65.47 \\
Student est. & 81.93 & 59.93 & 64.67 & 54.89 & 78.53 & 55.34 & 65.88 \\
Teacher-student est. & 79.03 & 58.70 & 65.05 & 53.54 & 77.90 & 52.83 & 64.51 \\
\bottomrule
\end{tabular}

\caption{Ablation study for importance weight estimation on Llama3.1-8B}
\label{table:weighting}
\end{threeparttable}
\end{table*}

\begin{table*}[t]
\centering
\begin{threeparttable}
\centering

\definecolor{lightgray}{gray}{0.9}

\begin{tabular}{lccccccc}
\toprule
\textbf{Method} & \textbf{HellaSwag} & \textbf{Arc} & \textbf{MMLU} & \textbf{TruthfulQA} & \textbf{Winogrande} & \textbf{GSM8k} & \textbf{Average} \\
\midrule
\textbf{Origin} & 82.25 & 63.92 & 66.65 & 55.22 & 79.29 & 57.47 & 67.42 \\
\midrule
Using student & 81.16 & 59.76 & 65.24 & 54.70 & 78.06 & 52.69 & 65.27 \\
\bottomrule
\end{tabular}

\caption{Ablation study for reference model on Llama3.1-8B}
\label{table:student_ref}
\end{threeparttable}
\end{table*}

\subsection{Importance weight estimation}
To calculate weights, we adapt the methodology from TIS-DPO~\cite{liu2024tisdpo}, which uses a pair of contrastive language models to estimate rewards. In our CTPD framework, we leverage the superior capabilities of the teacher model to construct this contrastive pair, thereby enhancing the guidance provided by the weights.

Specifically, we designate a standard DPO-trained version of the teacher model as the positive model, \(\pi^{+}\), and a reverse DPO-trained version as the negative model, \(\pi^{-}\). The importance weight \(w_t\) for each aligned span \(p^t\) is then estimated using the log-probability ratio between this contrastive pair:

\begin{equation*}
   w_t = k \cdot \exp(\mu \cdot \text{clamp}(\log\frac{\pi^{+}(p^t \mid x, p^{<t})}{\pi^{-}(p^t \mid x, p^{<t})}, L, U))
   \label{eq:w_t_estimate}
\end{equation*}

Here, the clamp limits \(L\) and \(U\) are used to stabilize the optimization process by reducing variance.
The teacher model's advanced capabilities enable it to capture nuanced differences, creating effective contrastive LLM pairs. By using this expert model to generate the contrastive signals that form our weights, we effectively distill its fine-grained reward judgments onto the student model, guiding the optimization process more effectively.

\section{Experiments}
Our comprehensive experiments show that our proposed CTPD method consistently outperforms existing techniques in alignment and distillation across various benchmarks. Furthermore, our ablation studies demonstrate that using a teacher model to determine the importance of aligned spans is a significantly more effective weighting strategy.
\subsection{Settings}
\label{sec:exp:setup}
\subsubsection{Baselines and LLMs.}
We evaluate CTPD across two scales: a small-scale pair using Qwen 2.5 7B as the teacher and Llama 3.2 1B as the student, and a large-scale pair with Qwen 2.5 14B as the teacher and Llama 3.1 8B as the student. We benchmark against two baseline categories: \textbf{preference alignment methods}---DPO~\cite{rafailov2023direct}, which directly optimizes the log-odds of preferred over rejected responses without an explicit reward model, and TIS-DPO~\cite{liu2024tisdpo}, which adds token-level importance weights so updates focus on high-reward parts of the answer---and \textbf{cross-tokenizer knowledge distillation methods}---ULD~\cite{boizard2024towards}, which aligns teacher and student logits under mismatched vocabularies via a Wasserstein distance; DSKD~\cite{zhang2024dual}, which projects representations into each other's spaces with a shared prediction head; and Multi-Level OT~\cite{cui2025multi}, which uses optimal transport at both token and sequence levels to preserve local and global logit structure during distillation.

\subsubsection{Datasets and Evaluation Metrics.}
For fine-tuning, we utilize the \textbf{UltraFeedback Binarized} dataset, available through Hugging Face\footnote{\url{https://huggingface.co/datasets/HuggingFaceH4/ultrafeedback_binarized}}, which contains over 63k high quality preference pairs. To assess model performance, we adopt the methodology of the \textbf{HuggingFace Open LLM Leaderboard}~\cite{beeching2023openllm}, implemented via the Language Model Evaluation Harness~\cite{gao2024harness}. This framework provides a robust assessment across six established benchmarks targeting key LLM capabilities:
commonsense reasoning (ARC~\cite{clark2018think}, HellaSwag~\cite{zellers2019hellaswag}, and Winogrande~\cite{sakaguchi2021winogrande}), 
multi-task language understanding (MMLU~\cite{hendrycks2020mmlu}),
factual accuracy (TruthfulQA~\cite{lin2021truthfulqa}),
mathematical reasoning (GSM8k~\cite{cobbe2021gsm8k}).
Collectively, these benchmarks provide a rigorous and multifaceted framework for assessing both alignment quality and general model competence. 

\subsubsection{Hyperparameters.} We trained our models for one epoch in all stage using the AdamW optimizer~\cite{loshchilov2017decoupled} with a global batch size of 16 distributed across eight NVIDIA H100-80GB GPUs. A cosine learning rate scheduler with a 5\% warmup period was used for all training stages. The random seed is globally set to 0.

\begin{itemize}
    \item \textbf{SFT:} For the initial SFT of student and teacher models, we used a learning rate of \(4 \times 10^{-6}\).

    \item \textbf{Positive and Negative Teacher Training:} In the subsequent phase for training positive and negative teacher models, we lowered the learning rate to \(2 \times 10^{-6}\) and set the DPO loss hyperparameter \(\beta\) to 0.3.

    \item \textbf{CTPD:} For our proposed CTPD framework, the learning rate was \(1 \times 10^{-6}\) and \(\beta\) was 0.1. For our proposed weight estimation method, we set the scaling factor $k=1$ and clipped the importance weights to the range $[L, U] = [-0.5, 1.5]$. For positive and negative samples we set $\mu$ to 1 and -1, respectively.
\end{itemize}

\subsection{Main Results}
\label{exp_method}

\subsubsection{Comparison with Preference Alignment Baselines.}
As illustrated in Table \ref{tab:benchmark_results_booktabs}, when compared to preference alignment techniques, CTPD demonstrates superior average performance in both cases, outperforming a strong TIS-DPO baseline by significant margins of +1.26 and +0.66 points, respectively. The improvements are consistent across individual tasks, with notable gains on GSM8k (+3.16 over TIS-DPO) and TruthfulQA (+2.85 over TIS-DPO). These benchmarks require a high degree of reasoning and factual precision, highlight the strength of our approach.

\subsubsection{Comparison with Knowledge Distillation Baselines.}
The results in Table~\ref{tab:benchmark_results_booktabs} also highlight that CTPD is a more effective method for leveraging teacher models than traditional knowledge distillation (KD), which primarily relies on an alignment of logits or intermediate representations. The consistent performance improvements across all benchmarks underscore the flexibility and robustness of our method. These findings suggest a promising new direction for knowledge distillation research.

\subsection{Ablation study}
\label{sec:exp:ablation}

\subsubsection{Influence of Different Weighting Strategies.}
To investigate the influence of various weighting strategies on the performance of CTPD, we conducted a comprehensive ablation study. We experimented with several distinct approaches:
\begin{itemize}
    \item \textbf{Random Weight:} Weights were uniformly sampled from the range of $(-1, 1)$.
    \item \textbf{Average Weight:} The original weight of each aligned span in our method were divided by the length of the span.
    \item \textbf{Student Estimate:} We employed two contrastive student models to estimate the weights.
    \item \textbf{Teacher-Student Estimate:} We utilized SFT checkpoints of both teacher and student models as positive and negative models to estimate the weights.
\end{itemize}

As illustrated in Table~\ref{table:weighting}, our proposed method (Origin) consistently achieves the best performance across all benchmarks. The \textit{Average}, \textit{Student Estimate}, and \textit{Teacher-Student Estimate} strategies all yield reasonable performance but are clearly surpassed by our approach. In contrast, the \textit{Random Weight} strategy leads to a substantial degradation in performance. These results underscore the critical importance of accurate weight estimation. Furthermore, they confirm the advantage of employing a teacher-guided approach, as implemented in CTPD, for achieving superior performance.

\subsubsection{Using the Student Model as the Reference Model.}
We also explored how the choice of the reference model affects CTPD's performance. For this analysis, we used the student model as the reference, instead of the teacher model typically employed in our CTPD framework. The results, presented in Table~\ref{table:student_ref}, show that using the teacher model to guide the policy achieves superior performance. This outcome validates our approach, underscoring the effectiveness of using a stronger model as a reference to direct the policy model's learning process.

\section{Conclusion}
In this work, we introduced Cross-Tokenizer Preference Distillation (CTPD), the first unified framework designed to transfer human-aligned behavior from a large teacher model to a smaller student model, even in the presence of heterogeneous tokenizers. By leveraging \textbf{Aligned Span Projection} mechanism that operates on character-level intervals, CTPD effectively bridges the gap between incompatible token spaces. We further developed a cross-tokenizer extension of TIS-DPO and Teacher-Anchored Reference approach to enable fine-grained, white-box distillation of preference signals. Extensive experiments demonstrate the effectiveness of our approach in advancing the state-of-the-art, overcome the problem of cross-tokenizer in preference distillation. Future work could explore extending CTPD to other forms of knowledge transfer, such as distilling specific skills or factual knowledge, and investigating its applicability in even more resource-constrained environments.

\section*{Acknowledgments}
Linh Ngo Van is funded by Vietnam National Foundation for Science and Technology Development (NAFOSTED) under grant number 102.05-2025.16. Ngan Nguyen was supported by FPT Smart Cloud, which contributed significantly to the completion of this work. Trung Le was supported by the Air Force Office of Scientific Research under award number FA9550-23-S-0001.

\bibliography{aaai2026}

\end{document}